\documentclass[letterpaper]{article} 
\usepackage[margin=0.75in]{geometry}
\usepackage{newtxtext}
\AtBeginDocument{\twocolumn}
\usepackage[hyphens]{url}
\usepackage{graphicx}
\usepackage{natbib}
\usepackage{caption}
\usepackage{booktabs}
\usepackage{colortbl}
\usepackage{amsfonts}
\usepackage{amssymb}
\usepackage{nicefrac}
\usepackage{amsmath}
\usepackage{multirow}
\usepackage{siunitx}
\usepackage[hidelinks]{hyperref}
\hypersetup{
  pdftitle={VoLN: Vision-Only Long-Horizon Navigation---Paradigm, Benchmark, and Method},
  pdfauthor={Jiabin Lou, Haopeng Wang, Yuanshuai Wang, Xinyu Liu, Xuxin Lv, Yuxin Guo, Lei Huang, Rongye Shi, and Wenjun Wu}
}
\newcommand{\best}[1]{{\bfseries #1}}
\definecolor{volnrow}{RGB}{210,220,248}

\title{VoLN: Vision-Only Long-Horizon Navigation---Paradigm, Benchmark, and Method}
\author{
    Jiabin Lou\textsuperscript{1,2},
    Haopeng Wang\textsuperscript{1,2},
    Yuanshuai Wang\textsuperscript{1},
    Xinyu Liu\textsuperscript{1},
    Xuxin Lv\textsuperscript{1},\\[0.2em]
    Yuxin Guo\textsuperscript{1},
    Lei Huang\textsuperscript{1},
    Rongye Shi\textsuperscript{1,2},
    and Wenjun Wu\textsuperscript{1,2,*}\\[0.45em]
    \small \textsuperscript{1}Beihang University, Beijing 100191, China\\
    \small \textsuperscript{2}Hangzhou International Innovation Institute, Beihang University, Hangzhou 311115, China\\
    \small \textsuperscript{*}Corresponding author: Wenjun Wu\\[-0.1em]
    \small \nolinkurl{loujiabin@buaa.edu.cn}; \nolinkurl{wwj09315@buaa.edu.cn}
}
\date{}

\begin{document}

\clubpenalty=0
\widowpenalty=0

\maketitle
\begingroup
\renewcommand{\thefootnote}{}
\footnotetext{This work was supported by the National Key Research and Development Program of China under Grant 2025YFF1505704.}
\endgroup

\begin{abstract}
Vision-and-Language Navigation (VLN) enables embodied agents to follow natural-language instructions. However, route-level instructions commonly encode spatial priors, such as orientation, distance, and layout, that are not explicitly available from onboard sensing at deployment in open, GPS-denied environments. Benchmark performance under such interfaces therefore jointly reflects visual navigation ability and the use of route structure explicitly supplied by the task description. As a complementary formulation, we propose Vision-Only Long-Horizon Navigation (VoLN), which shifts route-relevant information from externally supplied instructions and global guidance to locally observable in-scene cues. In VoLN, goal views specify the destination, while route-relevant information is available only through locally observable in-scene cues that the agent must detect, interpret, and select online. We instantiate VoLN for aerial navigation through VoLN-UAV, a 7,210-episode benchmark that combines long-horizon goal-directed flight, continuous 3D motion, large viewpoint changes, and context-dependent beacon selection. We further provide VoLN-MLLM as an initial reference baseline. It aligns self-supervised visual features with a structured semantic space and predicts short-horizon waypoint segments from observation history, goal views, retrieved visual--semantic tokens, and proprioception. On the five-environment Test-Unseen split, it obtains success rates of 7.4\%, 4.5\%, and 1.8\% on Easy, Normal, and Hard episodes, respectively. These results provide an initial evaluation of VoLN and reveal substantial remaining challenges in long-horizon evidence integration, cross-view goal matching, and closed-loop stability.
Project page: \url{https://admire-ljb.github.io/VoLN-UAV/}.
\end{abstract}

\section{Introduction}

\begin{figure}[t]
  \vspace*{\dimexpr1.5\baselineskip\relax}
  \centering
  \includegraphics[width=\linewidth]{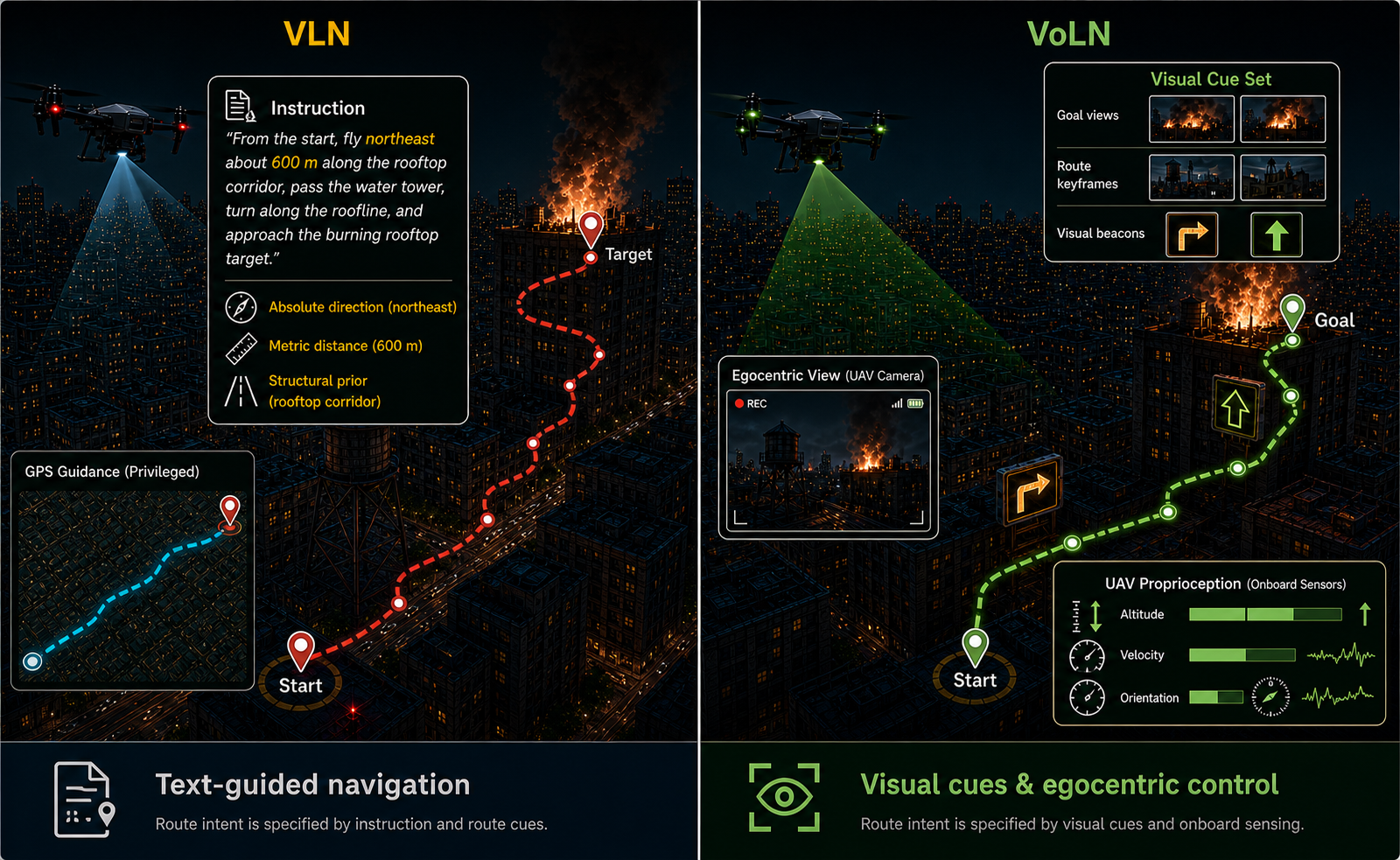}
  \caption{The instruction-based setting exposes route-level information through language and global guidance, whereas VoLN specifies the destination visually and presents route-relevant cues within the observable scene.}
  \label{fig:voln}
\end{figure}

Vision-and-Language Navigation (VLN) maps high-level semantic instructions into physical actions. Much of its early progress centered on indoor, ground-level agents operating in bounded spaces that are readily described in language and represented by maps or topological graphs. Applying the same instruction interface to aerial agents introduces a different operating regime: navigation unfolds in open 3D space, destinations frequently lie outside the current field of view, and long-range trajectories involve substantial changes in position, altitude, and viewpoint. Route instructions in such settings are commonly authored from trajectories planned with global scene knowledge and encode absolute orientation, metric distance, or route structure. These quantities provide an effective means of specifying a route, but they are not directly observed through onboard sensing at the corresponding decision points. Benchmark performance therefore reflects a combination of visual perception, language grounding, and the use of route structure conveyed by the instruction, making their respective contributions difficult to disentangle.

To address this issue, we introduce Vision-Only Long-Horizon Navigation (VoLN). During execution, VoLN removes externally supplied task-level route instructions and global navigation signals, including GPS, global maps, and shortest-path annotations, from the policy interface. As illustrated in Fig.~\ref{fig:voln}, goal views specify the destination, whereas route-relevant information is encountered only as locally observable in-scene cues, including semantic beacons. The agent must detect these cues from egocentric observations, interpret their meaning, and select those relevant to the current task online, while proprioception provides onboard motion state.

We instantiate VoLN in aerial navigation, where an Unmanned Aerial Vehicle (UAV) operates in continuous 3D space under substantial viewpoint and scale variation and flight-dynamics constraints. This setting stresses cross-view re-identification, cue selection, and closed-loop control simultaneously. Our benchmark, VoLN-UAV, spans diverse simulated environments and embeds the cue-discrimination problem in the scene itself: active beacons provide route-relevant guidance, while passive beacons with similar visual forms provide structured distractors. Evaluation measures goal convergence, trajectory quality, and closed-loop reliability.

To provide an initial solution, we introduce VoLN-MLLM, a two-stage visual--semantic planning framework. The first stage aligns self-supervised visual features with a structured semantic space, providing comparable representations for observations, goal views, and visible scene cues. The second stage integrates the aligned visual evidence with goal views and proprioception to generate short-horizon UAV trajectories in closed loop. The resulting experiments provide an initial benchmark reference and highlight recurring challenges under viewpoint change, visually similar distractors, and long-horizon execution.

Our contributions are:
\begin{itemize}
    \item \textbf{Task formulation.} We formulate VoLN as a long-horizon navigation paradigm in which goal views specify the destination, while the agent infers route-relevant information online from locally observable scene cues.
    \item \textbf{Benchmark.} We introduce VoLN-UAV, a 7{,}210-episode benchmark for long-horizon aerial navigation in continuous 3D environments, featuring active and passive semantic beacons and dedicated evaluation splits for seen and unseen environments.
    \item \textbf{Method.} We present VoLN-MLLM, a two-stage visual--semantic planning framework that first aligns observations and goal views with a structured semantic space and then generates short-horizon trajectories through cue-conditioned closed-loop planning.
\end{itemize}

\section{Related Work}

\subsection{Navigation task interfaces}

A navigation benchmark is shaped by its task interface, the channel through which intent reaches the agent. Language remains the dominant choice in VLN. NavGPT exemplifies explicit language-model reasoning for sequential action prediction \cite{zhou2024navgpt}. In aerial multi-agent autonomy, TALKER uses language task descriptions to activate and plan over a learned action-primitive library \cite{lou2025talker}. MapGPT adds an explicit topological memory for long-horizon planning \cite{chen2024mapgpt}. Reinforcement post-training for continuous control is explored in VLN-R1 \cite{qi2025vlnr1}. VLNVerse provides systematic evaluation across models and datasets \cite{lin2025vlnverse}. NavFoM learns transferable navigation priors from large-scale language supervision \cite{zhang2025navfom}.

Under this interface, benchmark performance jointly reflects perception, language grounding, and route-level information expressed in the instruction.
A complementary line of work specifies the navigation goal visually. End-to-end policies learn cross-view correspondence for image-goal reaching \cite{bono2024end2endimagenav}, transformer architectures strengthen sequential decision making \cite{pelluri2024transformersignav}, GaussNav grounds the goal in an explicit 3D Gaussian scene representation \cite{lei2024gaussnav}, IGL-Nav performs incremental 3D Gaussian localization \cite{guo2025iglnav}, and NavigateDiff introduces diffusion-based prediction \cite{qin2025navigatediff}. These methods primarily study terminal visual-goal grounding. Long-horizon navigation in which route-relevant information must be detected, interpreted, and selected online from locally observable in-scene cues remains comparatively less explored.

\subsection{Aerial navigation}
Aerial navigation research spans two related levels: motion planning and control in open 3D space, and semantic task execution under continuous flight. At the motion level, optimization-based methods explicitly model terrain, obstacle, and flight constraints. HHPSO uses heuristic hybrid particle swarm optimization for real-time quadcopter path planning and validates the resulting trajectories in simulation and real-flight experiments \cite{lou2024hhpso}. Learning-based control provides a complementary direction. Swift combines simulation-trained deep reinforcement learning with onboard sensing for agile real-world flight \cite{kaufmann2023swift}. Air Learning provides an open simulation and gym environment for deep reinforcement learning in resource-constrained visual UAV navigation \cite{krishnan2021airlearning}. Air-M further provides a visual-reality many-agent reinforcement learning platform for large-scale training and sim-to-real evaluation of aerial systems \cite{lou2023airm}.

At the task level, aerial VLN studies how UAVs interpret semantic instructions and execute them through onboard perception and control. AerialVLN provides an early city-scale formulation and data-construction pipeline \cite{liu2023aerialvln}; OpenUAV emphasizes high-fidelity flight control and assistant-guided evaluation \cite{wang2024openuav}; and OpenFly scales the collection of outdoor instruction--trajectory data \cite{gao2025openfly}. Corresponding methods include the end-to-end multimodal policy of UAV-VLN \cite{saxena2025uavvln}, the hierarchical planning and global memory of CityNavAgent \cite{zhang2025citynavagent}, and the staged training and interpretable reasoning of FlightGPT \cite{cai2025flightgpt}.

Overall, aerial navigation has advanced substantially. Existing aerial VLN benchmarks, however, commonly provide route information explicitly through natural-language instructions or other task-level guidance. Long-horizon aerial navigation in open 3D environments, with locally observable in-scene cues serving as en-route guidance, has received limited attention.

\subsection{Visual--semantic alignment and planning}
Acting on a visually specified goal requires semantic grounding, memory, and foresight. Pretrained vision--language models map observations into shared semantic spaces that support planning: VLFM constructs vision--language value maps for zero-shot target search \cite{yokoyama2024vlfm}, PixelNav specifies targets directly in pixel space \cite{zhang2024pixelnav}, and Find Everything balances multiple targets through score-map inference \cite{choi2024findeverything}. To maintain evidence over long horizons, Tag Map stores explicit text-based maps \cite{zhang2024tagmap}, E2Map updates maps from experience \cite{kim2024e2map}, and ReMEmbR retrieves from spatio-temporal memory \cite{anwar2025remembr}. Predictive methods add foresight: Imagine-Before-Go completes unseen semantic regions \cite{Zhang_2024_CVPR}, while WMNav \cite{nie2025wmnav}, ForesightNav \cite{shah2025foresightnav}, and VISTA \cite{huang2025vista} plan over imagined futures. At the system level, AERIS coordinates language-model-based planning and control at runtime \cite{lou2026aeris}. 

These studies provide useful foundations for semantic grounding, memory, and predictive planning. However, long-horizon closed-loop navigation remains less explored when the destination is specified visually and route-relevant information must be recovered from locally observable in-scene cues.

\section{The VoLN Paradigm}
\label{sec:problem}
\begin{figure}[h]
  \centering
  \includegraphics[width=0.95\linewidth]{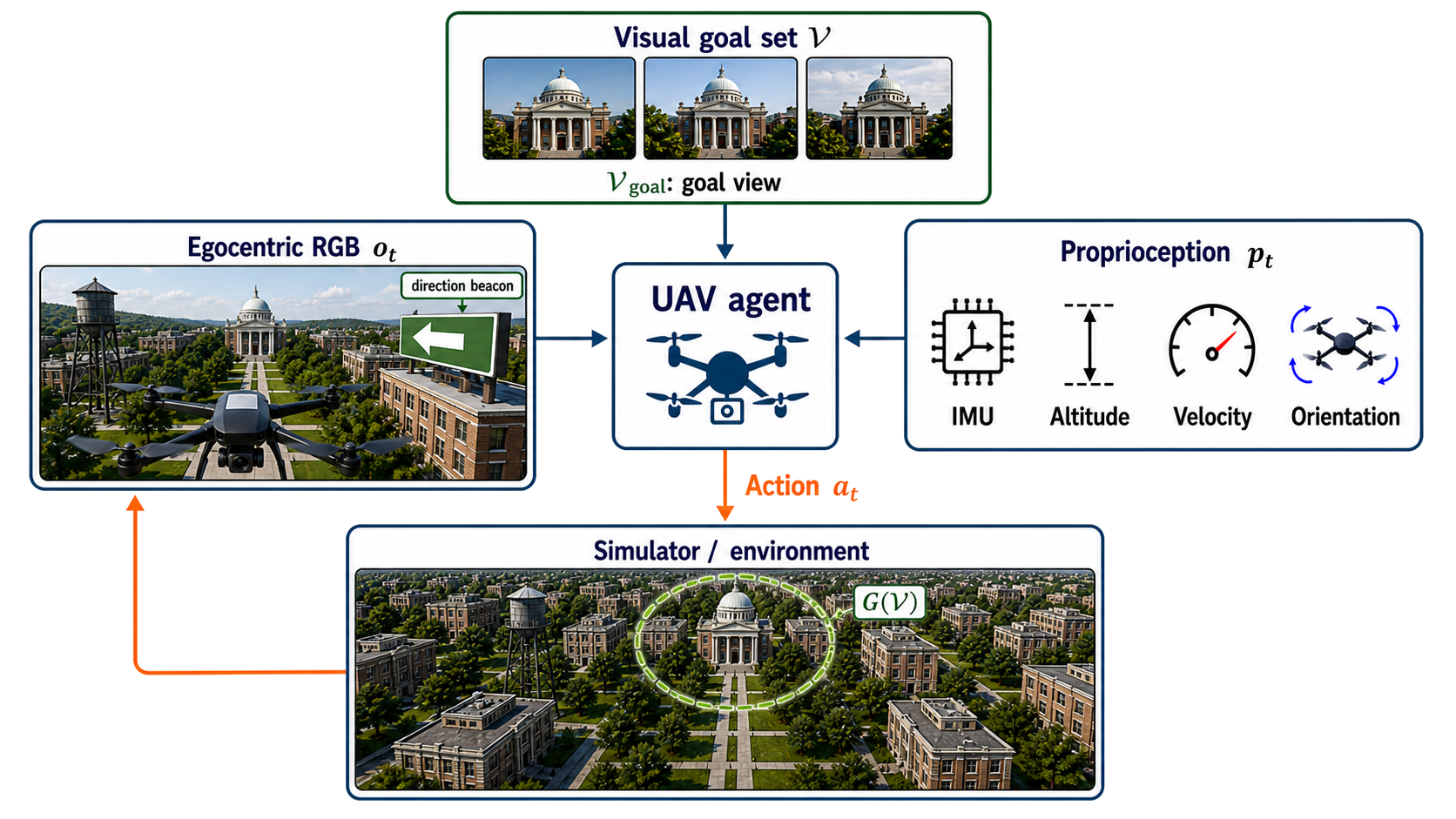}
  \caption{The VoLN interaction. The policy maps goal views $\mathcal{V}$, observations $o_t$, and proprioception $p_t$ to closed-loop actions $a_t$; $G(\mathcal{V})$ serves only for evaluation.}
  \label{fig:voln_setup}
\end{figure}

We formulate VoLN as a goal-directed, long-horizon embodied navigation paradigm. At execution time, the policy receives no externally supplied task-level route instructions or global navigation signals. Instead, each episode provides a visual goal set $\mathcal{V}$ composed of images captured near the destination, while route-relevant information is available only through locally observable in-scene cues encountered through $o_t$ (Fig.~\ref{fig:voln_setup}). The agent must ground the goal views across changes in viewpoint and detect, interpret, and select relevant cues online during closed-loop interaction.

We model each episode as a partially observable sequential decision process with latent state $s_t \in \mathcal{S}$. At time step $t$, the agent receives an observation $x_t = (o_t, p_t)$, where $o_t$ denotes the egocentric RGB observation and $p_t$ denotes proprioception, implemented as platform-provided onboard signals such as IMU measurements, altitude, velocity, and orientation; GPS and world-frame position are excluded. The agent outputs an action $a_t \in \mathcal{A}$, which may be continuous or discrete and includes an explicit stop decision, and the environment evolves according to the transition kernel $P$ and observation function $\Omega$:
\begin{equation}
s_{t+1} \sim P(\cdot \mid s_t, a_t), \qquad x_{t+1} \sim \Omega(\cdot \mid s_{t+1}).
\end{equation}
The policy conditions on the interaction history $h_t = (x_0, a_0, \ldots, x_t)$ and selects actions according to
\begin{equation}
a_t \sim \pi(\cdot \mid h_t, \mathcal{V}).
\end{equation}

VoLN targets episodes in which the destination remains outside the current view over substantial portions of the trajectory and route-relevant cues are encountered at multiple decision points. The policy therefore uses $h_t$ to integrate evidence across the trajectory. As the agent moves, $o_t$ reveals semantic beacons and naturally occurring landmarks at different decision points. The policy interprets these observations in relation to $\mathcal{V}$ and the accumulated interaction context, selecting cues that are relevant to the current task. For evaluation, each task instance associates $\mathcal{V}$ with a goal region $G(\mathcal{V}) \subset \mathcal{S}$. Starting from a designated initial pose, the agent has at most $T$ steps and succeeds by issuing the stop action inside $G(\mathcal{V})$.

\section{The VoLN-UAV Benchmark}
\label{sec:voln_uav_benchmark}
\subsection{Simulation Environments}
\begin{figure}[h]
  \centering
  \includegraphics[width=0.95\linewidth]{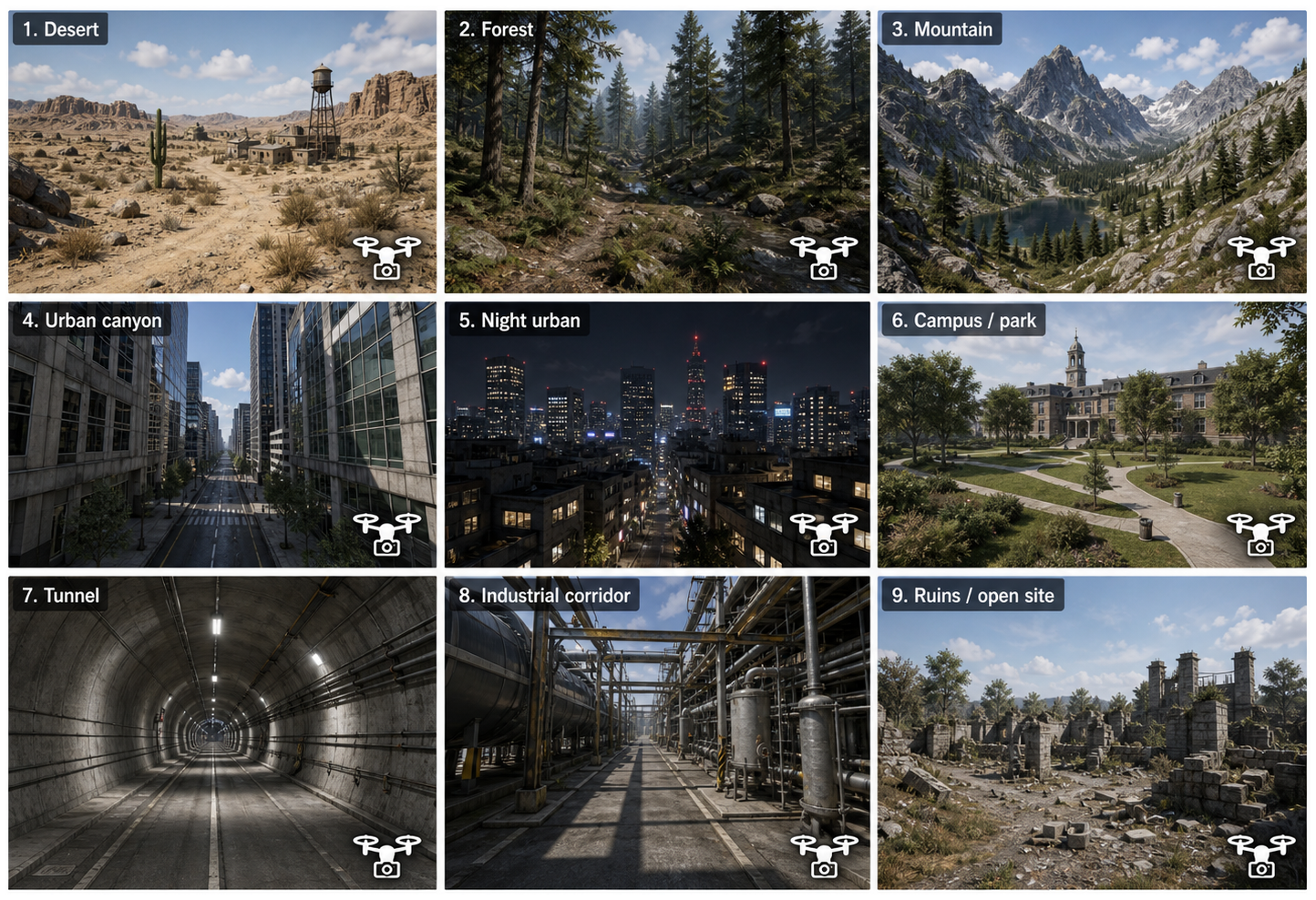}
  \caption{Representative VoLN-UAV environments.}
  \label{fig:voln_uav_environment}
\end{figure}

VoLN-UAV is built with Unreal Engine and Microsoft AirSim, which provide high-fidelity rendering and UAV simulation across the benchmark environments.
The environment pool contains 17 distinct environments drawn from selected scenes adapted from existing open-source aerial VLN benchmarks, including AerialVLN~\cite{liu2023aerialvln} and OpenUAV~\cite{wang2024openuav}, together with additional custom-built environments, $\mathcal{E} = \mathcal{E}^{\mathrm{open}} \cup \mathcal{E}^{\mathrm{custom}}$. This hybrid design preserves compatibility with existing aerial benchmarks while expanding scene diversity. As shown in Fig.~\ref{fig:voln_uav_environment}, the benchmark spans natural and built environments, from deserts, forests, and mountains to urban canyons, tunnels, and industrial corridors, with substantial variation in layout, visibility, altitude change, and landmark density.

\subsection{Benchmark Construction Pipeline}
\begin{figure*}[t]
    \centering
    \includegraphics[width=0.95\linewidth]{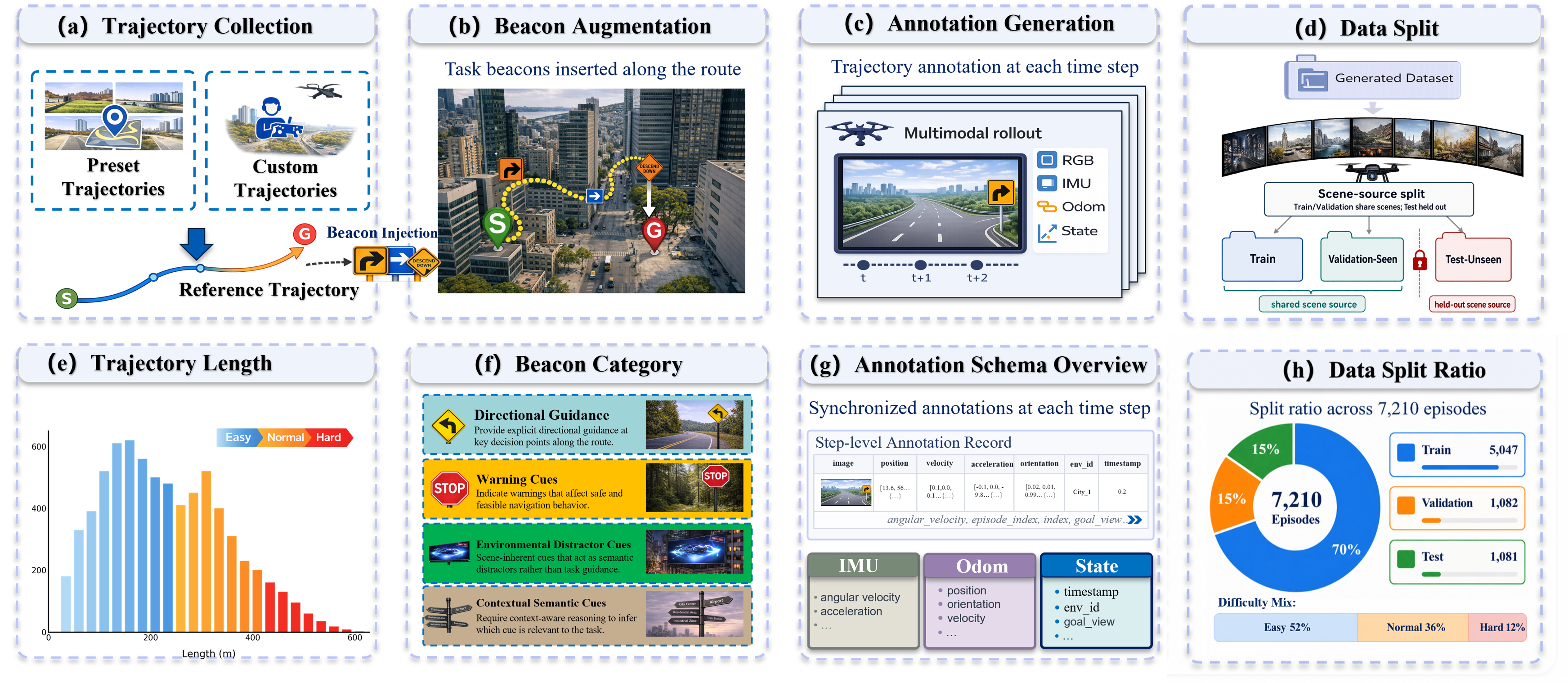}
    \caption{Overview of VoLN-UAV. Panels (a)--(d) show trajectory collection, beacon augmentation, annotation generation, and scene-source splitting; panels (e)--(h) summarize trajectory length, beacon categories, annotation fields, and split statistics.}
    \label{fig:voln_uav_pipeline}
\end{figure*}

We instantiate VoLN-UAV through the trajectory-centric pipeline illustrated in Fig.~\ref{fig:voln_uav_pipeline}. The following paragraphs detail its construction stages and resulting data organization.

\paragraph{Trajectory collection.}
Panels (a) and (e) relate reference-route collection to the resulting path-length distribution. For each episode, we sample a scene from $\mathcal{E}$ and select a reference route from the corresponding route pool. The pool contains predefined routes from existing datasets and custom routes collected by trained human operators following a standardized recording protocol. The predefined trajectories follow the route and goal settings of the source datasets, while the custom trajectories expand coverage of underrepresented flight patterns, including long corridors, sharp turns, altitude transitions, and ambiguous junctions. VoLN-UAV stratifies episodes by accumulated reference-path length, $L_{\mathrm{ref}}(\xi)=\sum_{t=1}^{T}\|\mathbf{r}_t-\mathbf{r}_{t-1}\|_2$, where $\mathbf{r}_t$ denotes the reference position at step $t$. Episodes are labeled \textbf{Easy} ($L_{\mathrm{ref}}<L_1$), \textbf{Normal} ($L_1\leq L_{\mathrm{ref}}<L_2$), or \textbf{Hard} ($L_{\mathrm{ref}}\geq L_2$), with $L_1=300\,\mathrm{m}$ and $L_2=450\,\mathrm{m}$ across all splits.
\paragraph{Beacon augmentation and cue categories.}
Panels (b) and (f) show beacon placement and the corresponding cue categories. Each reference trajectory is augmented with three to five active beacons, sparsely placed at decision points as task-relevant cues. Each environment is additionally populated with approximately 150 passive beacons that remain fixed across episodes and provide semantic clutter. The beacons span four semantic categories: directional guidance, warning cues related to feasible flight, environmental distractors, and contextual cues whose relevance depends on the current task. During execution, the UAV accesses beacons only through its egocentric observation $o_t$, which may contain both task-relevant active beacons and passive beacons present in the scene.

\paragraph{Multimodal rollout and annotation schema.}
Panels (c) and (g) present the synchronized multimodal rollout and its step-level annotation schema. For each reference trajectory $\xi=\{s_0,\ldots,s_T\}$, the simulator records synchronized egocentric RGB observations $\{o_0,\ldots,o_T\}$ and proprioceptive states $p_t$ at a fixed interval $\Delta t=2\,\mathrm{s}$. The final three RGB observations form the episode-level visual goal set:
\begin{equation}
\mathcal{V}(\xi)=\{o_{T-2},o_{T-1},o_T\}.
\end{equation}

For each step $t$, the observation and proprioceptive state are paired with the episode-level goal set $\mathcal{V}(\xi)$ to form a training sample, while the following $H$ states along the reference route define the short-horizon waypoint target $W_{t:t+H}$.

\paragraph{Dataset split and statistics.}
Panels (d) and (h) summarize the scene-source split and episode distribution. The dataset contains 7{,}210 episodes over 17 distinct environments: \textbf{Train} contains 5{,}047 episodes from 12 environments; \textbf{Validation-Seen} (VS) contains 1{,}082 episodes whose trajectories are disjoint from training but are drawn from 5 environments within the training pool; and \textbf{Test-Unseen} (TU) contains 1{,}081 episodes from 5 additional environments belonging to a held-out scene source. The splits correspond to an approximately 70\%/15\%/15\% episode ratio, and the aggregate difficulty mix is 52\% Easy, 36\% Normal, and 12\% Hard.
\section{The VoLN-MLLM Method}
\label{sec:voln_llm}

\begin{figure}[h]
    \centering
    \includegraphics[width=0.95\linewidth]{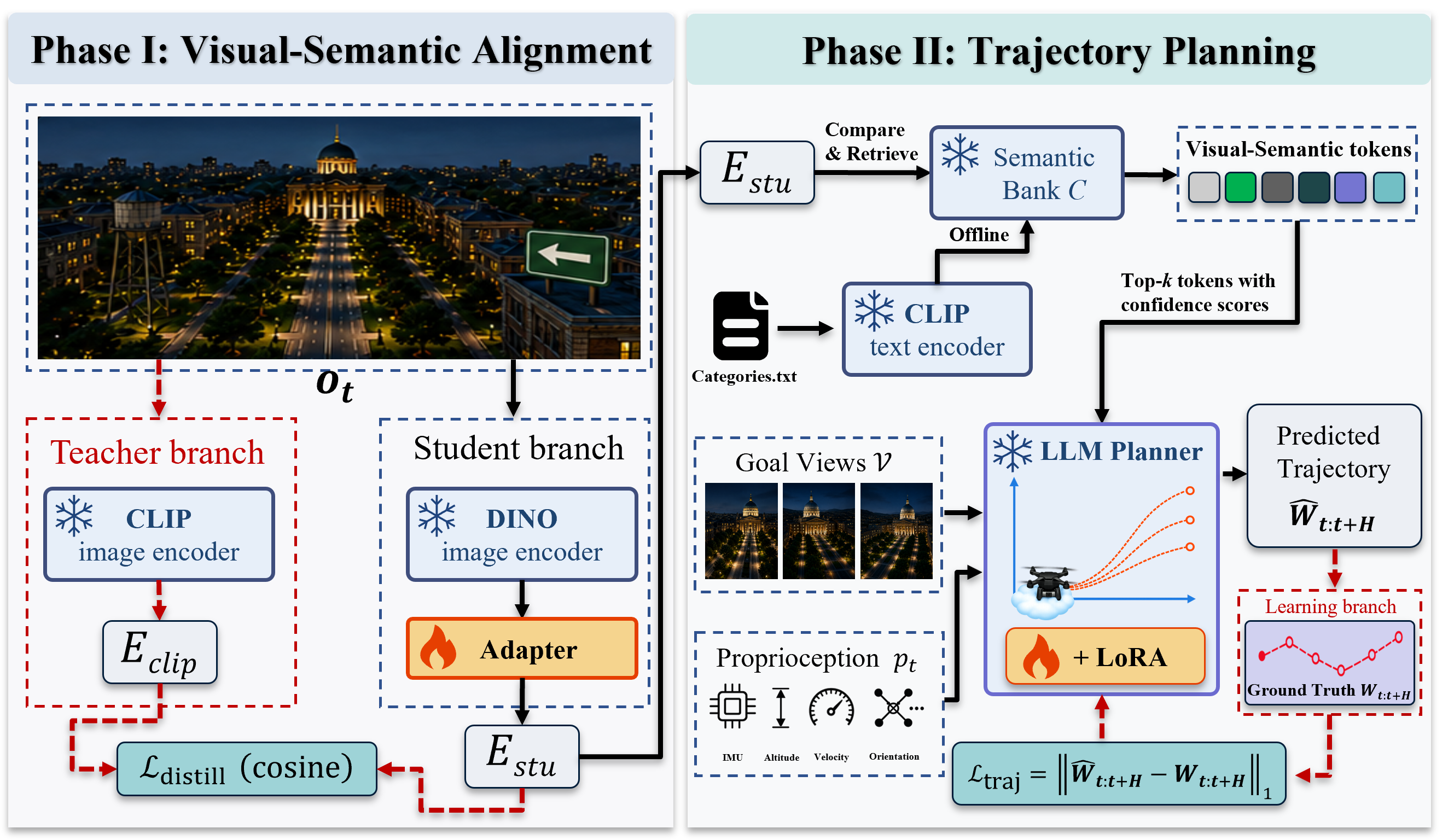}
    \caption{VoLN-MLLM overview. Phase I learns visual--semantic alignment; Phase II predicts short-horizon waypoints and stopping decisions. Dashed arrows indicate training branch; the stopping head is omitted for clarity.}
    \label{fig:algorithm}
\end{figure}

VoLN-UAV couples visual--semantic grounding with closed-loop trajectory generation: the agent must interpret locally observed route cues in relation to the goal views and convert this evidence into executable waypoint segments. We address these requirements with \textbf{VoLN-MLLM}, a two-stage visual--semantic planning framework (Fig.~\ref{fig:algorithm}). Phase I aligns DINO visual features with CLIP's joint image--text space, allowing observations to retrieve relevant concepts from a fixed semantic bank. Phase II employs a pretrained language-model planner. Aligned visual features from the recent observation history and goal views, retrieved visual--semantic tokens, and proprioception are projected into the planner's embedding space and jointly encoded for short-horizon waypoint and stopping prediction. The predicted segment is executed by a low-level flight controller, and the model replans from the subsequent observation. The semantic bank is constructed offline and shared across episodes, while execution follows the VoLN interface without externally supplied task-level route instructions or global navigation signals.

\paragraph{Visual--Semantic Alignment.}
Given an observation $o_t$, a frozen DINO backbone extracts a visual representation.
A lightweight trainable adapter, following the cross-space alignment principle of Talking to DINO~\cite{barsellotti2025talking}, maps this representation into the CLIP image-embedding space, producing a normalized student embedding $E_{\text{stu}}$.
During training of the adapter, a frozen CLIP image encoder provides the normalized teacher embedding $E_{\text{clip}}$ for the same image.
The adapter is optimized by a distillation objective:
\begin{equation}
\mathcal{L}_{\text{distill}} = \ell\!\left(E_{\text{stu}}, E_{\text{clip}}\right),
\end{equation}
where $\ell(\cdot,\cdot)$ is instantiated as cosine distance.
This stage updates only the adapter parameters and keeps the DINO backbone and CLIP encoders fixed.

\paragraph{Visual--semantic tokenization.} We construct a fixed semantic bank $\mathcal{C}$ whose entries are textual category descriptors encoded offline by a frozen CLIP text encoder. At each time step, the aligned visual embedding is compared with the bank embeddings using cosine similarity, and the top-$k$ entries are retained. Their category identifiers are encoded with the planner tokenizer, while their similarity scores are mapped to learned confidence embeddings. The resulting visual--semantic tokens are passed to the planner. 

\paragraph{Trajectory decoding.}
At each decision step, the aligned features of a fixed window of recent observations and the goal views are mapped to the planner's embedding dimension by a shared visual projector. A separate state projector maps proprioception $p_t$ to a state token. These embeddings are concatenated with the retrieved visual--semantic tokens and learned structural tokens marking the goal, history, semantics, state, and planning fields. The language-model planner jointly encodes this sequence, and the hidden state of the planning token is passed to a trajectory head and a binary stopping head. The two heads predict a short-horizon sequence of relative waypoints $\hat{W}_{t:t+H}$ in the current UAV body frame and stopping probability $\hat z_t$, respectively. The predicted sequence is tracked by the shared low-level flight controller, after which the next observation is acquired and the planner is invoked again.

\paragraph{Planner adaptation and supervision.}
We keep the pretrained language-model backbone frozen and adapt its attention and feed-forward projections using low-rank adaptation (LoRA)~\cite{hu2022lora}. The visual and state projectors, trajectory head, and stopping head are trained jointly with the LoRA parameters. At each step, the trajectory head predicts a segment of length $H$, supervised against demonstrations with an $\ell_1$ loss:
\begin{equation}
\mathcal{L}_{\text{traj}} = \left\|\hat{W}_{t:t+H} - W_{t:t+H}\right\|_1,
\end{equation}
where $\hat{W}_{t:t+H}$ denotes the predicted waypoint sequence and $W_{t:t+H}$ denotes the corresponding demonstrated sequence. Let $z_t\in\{0,1\}$ indicate whether the reference state lies inside the success region. The stopping head is trained with binary cross-entropy, and the complete objective is
\begin{equation}
\mathcal{L}=\mathcal{L}_{\text{traj}}+\lambda_{\text{stop}}\operatorname{BCE}(\hat z_t,z_t).
\end{equation}
At inference time, the policy stops when $\hat z_t$ exceeds a threshold $\tau$ selected on Validation-Seen; otherwise, it executes the predicted waypoint segment and replans.

\begin{table*}[h]
\centering
\footnotesize
\setlength{\tabcolsep}{2.6pt}
\renewcommand{\arraystretch}{1.05}
\caption{Results on Validation-Seen (VS) and Test-Unseen (TU) across difficulty levels.}
\label{tab:vln_main}

\resizebox{\textwidth}{!}{%
\begin{tabular}{@{}lc
S[table-format=3.1] S[table-format=3.1] S[table-format=3.1]
S[table-format=2.1] S[table-format=2.1] S[table-format=2.1]
S[table-format=2.1] S[table-format=2.1] S[table-format=2.1]
S[table-format=2.1] S[table-format=2.1] S[table-format=2.1]
S[table-format=2.1] S[table-format=2.1] S[table-format=2.1]
@{}}
\toprule
\multirow{2}{*}{Method} & \multirow{2}{*}{Split}
& \multicolumn{3}{c}{NE/m$\downarrow$}
& \multicolumn{3}{c}{SR/\%$\uparrow$}
& \multicolumn{3}{c}{OSR/\%$\uparrow$}
& \multicolumn{3}{c}{nDTW/\%$\uparrow$}
& \multicolumn{3}{c}{SPL/\%$\uparrow$} \\
\cmidrule(lr){3-5}\cmidrule(lr){6-8}\cmidrule(lr){9-11}\cmidrule(lr){12-14}\cmidrule(lr){15-17}
& &
\multicolumn{1}{c}{Easy} & \multicolumn{1}{c}{Normal} & \multicolumn{1}{c}{Hard}
& \multicolumn{1}{c}{Easy} & \multicolumn{1}{c}{Normal} & \multicolumn{1}{c}{Hard}
& \multicolumn{1}{c}{Easy} & \multicolumn{1}{c}{Normal} & \multicolumn{1}{c}{Hard}
& \multicolumn{1}{c}{Easy} & \multicolumn{1}{c}{Normal} & \multicolumn{1}{c}{Hard}
& \multicolumn{1}{c}{Easy} & \multicolumn{1}{c}{Normal} & \multicolumn{1}{c}{Hard} \\
\midrule


Random  & VS
& 268.5 & 308.7 & 398.9
& 0.6   & 0.0   & 0.0
& 1.8   & 0.8   & 0.2
& 28.3   & 20.0   & 12.1
& 0.4   & 0.0  & 0.0\\

Seq2Seq-VG & VS
& 205.8 & 251.6 & 307.4
& 1.2   & 0.5   & 0.1
& 5.4   & 2.9   & 1.0
& 30.1  & 21.0  & 11.8
& 0.9  & 0.3  & 0.1 \\

CMA-VG  & VS
& 170.2 & 211.9 & 261.3
& 1.9   & 0.9   & 0.2
& 7.6   & 4.3   & 1.9
& 34.5  & 25.2  & 16.1
& 1.3   & 0.6  & 0.1 \\

LAG-VG  & VS
& 118.7 & 154.9 & 203.6
& 2.8   & 1.5   & 0.5
& 7.8   & 4.6   & 1.9
& 29.7  & 20.4  & 12.7
& 1.9 & 1.0  & 0.3 \\
\rowcolor{volnrow}
\textbf{VoLN-MLLM} & VS
& \best{92.4}  & \best{126.8} & \best{171.5}
& \best{~8.7}   & \best{~5.4}   & \best{~~2.1}
& \best{13.4}  & \best{10.6}  & \best{~4.1}
& \best{54.8}  & \best{40.9}  & \best{25.7}
& \best{~6.5} & \best{~3.8} & \best{~~1.4} \\

\addlinespace
\midrule

Random  & TU
& 270.1 & 310.4 & 395.2
& 0.4   & 0.0   & 0.0
& 1.4   & 0.6   & 0.2
& 30.1   & 22.7  & 15.1
& 0.3   & 0.0   & 0.0 \\

Seq2Seq-VG & TU
& 208.6 & 254.8 & 309.9
& 1.0   & 0.4   & 0.1
& 4.8   & 2.5   & 0.9
& 28.9  & 21.4  & 13.0
& 0.7  & 0.3   & 0.0 \\

CMA-VG  & TU
& 174.5 & 216.8 & 266.1
& 1.6   & 0.8   & 0.2
& 6.5   & 3.9   & 1.7
& 33.2  & 26.4  & 18.5
& 1.1 & 0.6  & 0.1 \\

LAG-VG  & TU
& 122.4 & 158.3 & 206.7
& 2.3   & 1.2   & 0.4
& 6.4   & 3.8   & 1.7
& 28.1  & 20.5  & 14.0
& 1.5  & 0.7  & 0.2 \\
\rowcolor{volnrow}
\textbf{VoLN-MLLM} & TU
& \best{97.1}  & \best{131.4} & \best{176.8}
& \best{~7.4}   & \best{~4.5}   & \best{~~1.8}
& \best{14.6}  & \best{10.1}  & \best{~4.5}
& \best{53.1}  & \best{41.2}  & \best{28.0}
& \best{~5.7} & \best{~3.2} & \best{~~1.3} \\
\bottomrule
\end{tabular}%
}
\end{table*}
\section{Experiments}
\label{sec:experiments}

\subsection{Experimental Setup}

\paragraph{Implementation details.}
All experiments follow the VoLN-UAV protocol (Sec.~\ref{sec:voln_uav_benchmark}): at each decision step the agent receives only an egocentric RGB observation, proprioception, and the visual goal specification $\mathcal{V}$. World-frame poses are used only for supervision and evaluation and are not exposed to the policy.
The shared action interface is waypoint-based: at each decision step the policy emits a segment of $H=8$ relative three-dimensional waypoints together with a stop signal, a low-level controller tracks the segment, and an episode terminates when the policy stops or the step budget of $T=128$ decision steps is exhausted.
Our reference baseline, VoLN-MLLM~(Sec.~\ref{sec:voln_llm}), consists of a frozen DINOv3 ViT-B\/16 visual backbone, a lightweight adapter that aligns DINO features to the CLIP ViT-B/16 image-embedding space, and a frozen Vicuna-7B-v1.5 planning backbone~\cite{zheng2023judging}. Learned visual and proprioceptive projectors map the permitted task inputs to the Vicuna embedding dimension, and LoRA modules of rank 16 adapt its attention and feed-forward projections. A trajectory head predicts eight such waypoints, while a binary stopping head produces the stop signal.
All methods operate in closed loop and share this action interface and stopping criterion.

\paragraph{Baselines.}
We compare VoLN-MLLM with a random policy and three visual-goal (VG) variants of representative instruction-following architectures, replacing their language inputs with the VoLN interface. All learned methods receive the same observation history, goal views, proprioception, and semantic tokens retrieved from the frozen bank, and share the same visual encoder, waypoint action space, and training targets. \textbf{Random} samples feasible actions uniformly. \textbf{Seq2Seq-VG}, based on \citet{anderson2018}, uses a recurrent encoder--decoder over visual--proprioceptive history, the goal representation, and retrieved semantic tokens. \textbf{CMA-VG}, based on \citet{krantz2020beyond}, attends from the current visual state to history, goal, semantic tokens, and proprioception. \textbf{LAG-VG}, based on \citet{liu2023aerialvln}, separately attends to observation history and semantic tokens before waypoint prediction.

\paragraph{Evaluation metrics.}
Following common practice \cite{krantz2020beyond}, we report Success Rate (SR, stopping inside the $\epsilon=4$\,m goal region), Oracle Success Rate (OSR, entering the success region at any point), Navigation Error (NE, the final Euclidean distance to the goal), normalized Dynamic Time Warping (nDTW) \cite{ilharco2019dtw}, and Success weighted by Path Length (SPL)~\cite{anderson2018evaluation}:
\begin{equation}
\mathrm{SPL}=\frac{1}{N}\sum_{i=1}^{N} S_i \cdot \frac{l_i}{\max(p_i,l_i)}\,,
\end{equation}
where $S_i\in\{0,1\}$ indicates whether episode $i$ is successful, $l_i$ denotes the shortest-path distance from the start position to the goal (computed offline for evaluation only), and $p_i$ is the actual path length executed by the agent.

\begin{figure*}[t]
  \centering
  \includegraphics[width=0.97\textwidth]{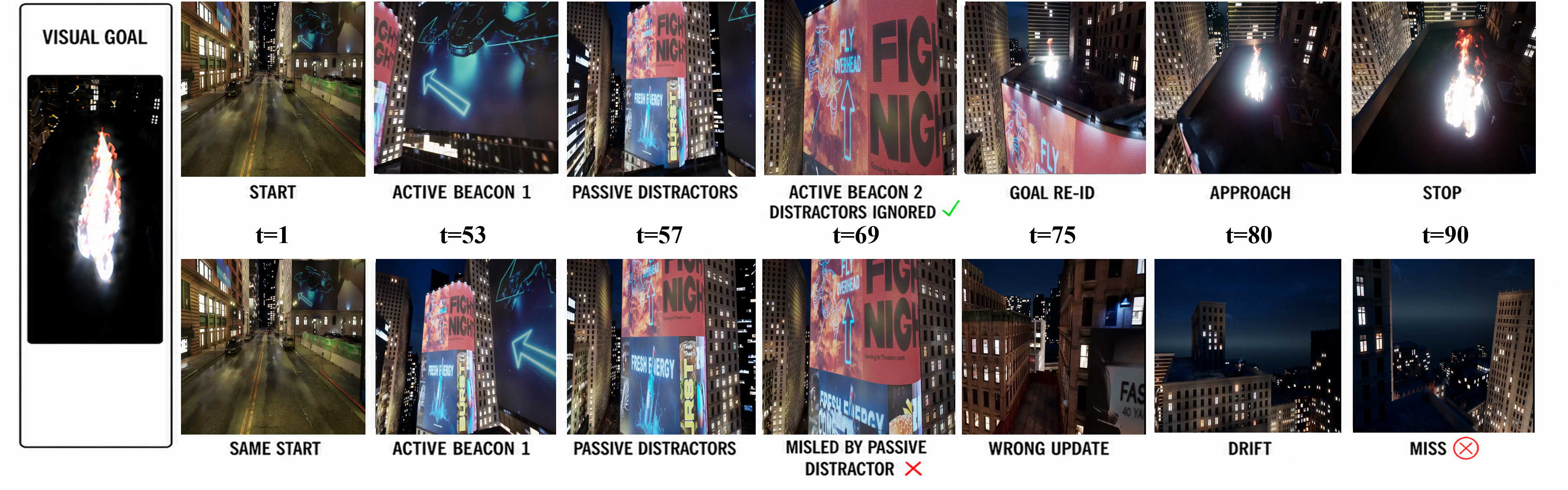}
  \caption{\textbf{Successful and failed rollouts.} Top: the trajectory passes the active-beacon locations and enters the goal region. Bottom: the trajectory diverges near the passive billboards and terminates outside the goal region.}
  \label{fig:qual_cases}
\end{figure*}

\subsection{Main Results}
\label{sec:exp_vln_main}

Table~\ref{tab:vln_main} compares VoLN-MLLM with the baselines on Validation-Seen and Test-Unseen across the Easy, Normal, and Hard subsets.

\paragraph{Overall performance.}
VoLN-MLLM yields the highest reported point estimate for every metric in Table~\ref{tab:vln_main}. On Test-Unseen, its SR reaches 7.4\%, 4.5\%, and 1.8\% across the three difficulty levels, compared with 2.3\%, 1.2\%, and 0.4\% for the strongest baseline, LAG-VG. Performance declines with increasing route difficulty for all methods, but the relative advantage of VoLN-MLLM persists. Nevertheless, the low absolute SR, particularly on the Hard subset, shows that long-horizon visual-only navigation remains challenging.

\paragraph{Trajectory quality and efficiency.}
Beyond terminal success, VoLN-MLLM has lower reported NE, indicating that its final positions are closer to the goal. Its higher reported nDTW corresponds to greater agreement between the executed and reference trajectories. These improvements are observed across both splits and all three path-length strata.
VoLN-MLLM also records the highest SPL, reflecting stronger combined performance in navigation success and path efficiency.

\begin{figure}[t]
  \centering
  \includegraphics[width=0.92\linewidth]{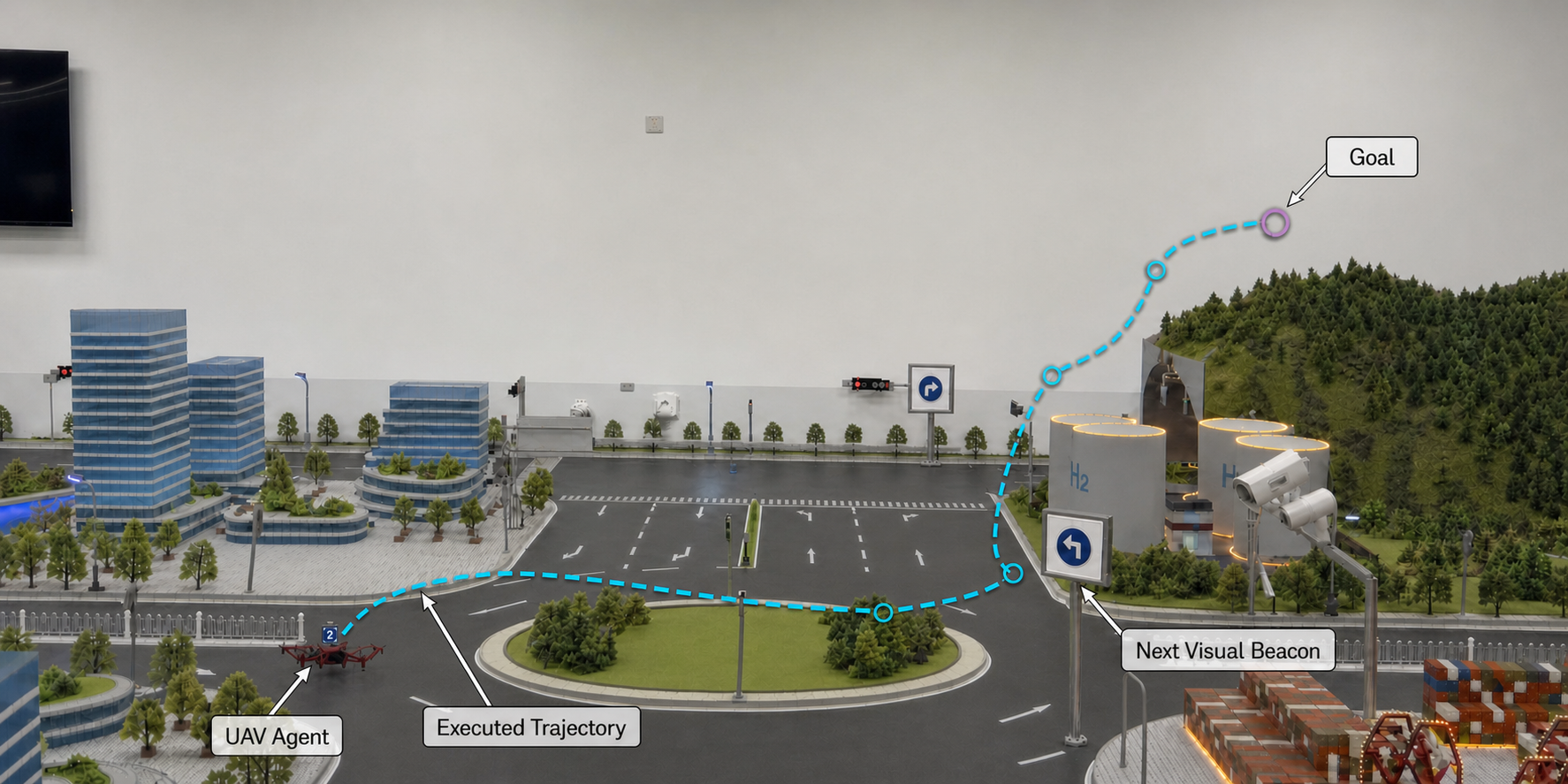}
  \caption{Physical testbed and a representative VoLN rollout.}
  \label{fig:real_testbed}
\end{figure}

\subsection{Supplementary Analyses}

In the ablation study, we additionally report cycle time (CT), measured from observation input to action output, and execution error rate (EER), defined as the percentage of planning cycles that exceed the time budget or produce invalid outputs.
\begin{table}[h]
  \centering
  \caption{Ablation results on the Test-Unseen split.}
  \label{tab:ablation_vln_tu}
  \setlength{\tabcolsep}{4pt}
  \renewcommand{\arraystretch}{1.05}
  \small
  \resizebox{\linewidth}{!}{
  \begin{tabular}{lccccccc}
    \toprule
    \textbf{Variant} &
    \textbf{CT (s)}$\downarrow$ &
    \textbf{EER (\%)}$\downarrow$ &
    \textbf{NE}$\downarrow$ &
    \textbf{SR}$\uparrow$ &
    \textbf{OSR}$\uparrow$ &
    \textbf{nDTW}$\uparrow$ &
    \textbf{SPL}$\uparrow$ \\
    \midrule
    VoLN-MLLM
      & 1.42 & \best{0.5} & \best{119.0} & \best{5.7} & \best{11.8} & \best{45.8} & \best{4.3} \\
    \midrule
    \ \ -- No-Align
      & \best{1.36} & 0.9 & 162.8 & 2.3 & 7.1 & 29.6 & 1.2 \\
    \ \ -- No-LoRA
      & 1.45 & 5.8 & 176.9 & 2.8 & 7.8 & 27.2 & 1.4 \\
    \ \ -- CLIP-Input
      & 1.98 & 1.5 & 158.6 & 2.9 & 8.2 & 30.9 & 1.6 \\
    \bottomrule
  \end{tabular}}
\end{table}
\paragraph{Ablation studies.}
On Test-Unseen, we ablate three components. {No-Align} retains the dimensional projection but removes CLIP-teacher supervision. {No-LoRA} freezes the planner backbone and removes the LoRA branch, while retaining the trained input projectors and prediction heads. {CLIP-Input} replaces the DINO backbone with CLIP image encoder. All variants use the same training split, seed pool, and evaluation budget. As shown in Table~\ref{tab:ablation_vln_tu}, No-Align causes the largest drop in SR ($5.7\rightarrow2.3$), while nDTW decreases from $45.8$ to $29.6$, highlighting the importance of compatibility between visual representations and the semantic space for effective grounding. Removing the LoRA branches yields the highest EER ($5.8\%$) and the lowest nDTW ($27.2$), suggesting that planner adaptation improves output reliability and trajectory fitting. CLIP-Input increases the cycle time from $1.42$\,s to $1.98$\,s and reduces SR to $2.9\%$. Overall, the results highlight the contributions of visual--semantic alignment to grounding quality, planner adaptation to output reliability and trajectory fitting, and robust visual representations to navigation success and efficient inference.
\paragraph{Success and failure cases.}
Fig.~\ref{fig:qual_cases} presents one successful and one failed navigation rollout under the same goal exemplar and initial context. The selected frames correspond to comparable stages of the two rollouts. In the successful case, the trajectory passes the active-beacon locations and eventually enters the goal region. In the failed case, the trajectory deviates near the passive billboards and terminates outside the goal region. Together, the two cases qualitatively illustrate trajectory divergence at a corresponding intermediate stage in a beacon-rich environment.

\paragraph{Physical testbed demonstration.}
We conduct a preliminary physical demonstration of the VoLN task interface in a controlled indoor testbed. The testbed comprises a scaled urban scene with roads, a roundabout, building clusters, vegetated terrain, and miniature directional beacons (Fig.~\ref{fig:real_testbed}). At the navigation-policy level, the UAV receives goal views, onboard RGB observations, and proprioception, following the same input interface used in simulation. In one representative rollout, the UAV traverses the roundabout and changes direction near the beacon-marked junction toward the hilltop goal region. This demonstration provides qualitative evidence that the VoLN task interface and closed-loop navigation pipeline can be instantiated on a controlled physical platform.

\section{Conclusion}
This work introduces VoLN, a vision-only long-horizon closed-loop navigation paradigm. During execution, VoLN removes externally supplied task-level route instructions and global navigation signals from the policy interface. Goal views specify the destination, while route-relevant information is available only through locally observable in-scene cues that the agent must detect, interpret, and select online. We instantiate this formulation in VoLN-UAV, a long-horizon aerial navigation benchmark featuring active and passive semantic beacons, continuous 3D control, and a scene-source-held-out test split.
VoLN-MLLM provides an initial reference baseline for this task. It maps self-supervised visual features into the semantic space defined by a fixed semantic bank and predicts short-horizon waypoint segments and stopping decisions from observation history, goal views, and proprioception. It produces the highest reported point estimates among the adapted baselines across both evaluation splits and all three difficulty levels. Nevertheless, its success rates on Test-Unseen remain 7.4\%, 4.5\%, and 1.8\% for Easy, Normal, and Hard episodes, respectively. The low success rates show that reliable long-horizon navigation remains unresolved under this interface. Progress requires integrating observational evidence over time, assessing the route relevance of in-scene cues, and limiting the accumulation of local errors during closed-loop execution. Finally, a controlled physical testbed demonstration provides a proof-of-concept instantiation of the VoLN interface beyond simulation.

\bibliographystyle{voln_natbib}
\bibliography{refs}          

\end{document}